\pdfoutput=1
\documentclass{article} 
\usepackage[preprint]{colm2025_conference}

\usepackage{microtype}
\usepackage{hyperref}
\usepackage{url}
\usepackage{booktabs}

\usepackage{lineno}

\usepackage{booktabs}
\usepackage{graphicx}
\usepackage{enumitem}
\usepackage{amsmath}
\usepackage{cleveref}
\usepackage{wrapfig} 

\usepackage{amsmath,amsfonts,bm}

\def\eqref#1{equation~\ref{#1}}

\def\1{\bm{1}}

\def\rvx{{\mathbf{x}}}

\DeclareMathAlphabet{\mathsfit}{\encodingdefault}{\sfdefault}{m}{sl}
\SetMathAlphabet{\mathsfit}{bold}{\encodingdefault}{\sfdefault}{bx}{n}

\definecolor{darkblue}{rgb}{0, 0, 0.5}
\hypersetup{colorlinks=true, citecolor=darkblue, linkcolor=darkblue, urlcolor=darkblue}

\title{Exploring Scaling Laws for EHR Foundation Models}

\author{\textbf{Sheng Zhang}$^{1}$, \textbf{Qin Liu}$^{2}$\thanks{Work completed as a student researcher at Microsoft Research.},~ \textbf{Naoto Usuyama}$^{1}$, \textbf{Cliff Wong}$^{1}$\\
\textbf{Tristan Naumann}$^{1}$, \textbf{Hoifung Poon}$^{1}$\\
\\
\textsuperscript{1} Microsoft Research\\
\textsuperscript{2} University of Southern California\\
}

\begin{document}

\ifcolmsubmission
\linenumbers
\fi

\maketitle

\begin{abstract}
    The emergence of scaling laws has profoundly shaped the development of large language models (LLMs), enabling predictable performance gains through systematic increases in model size, dataset volume, and compute. Yet, these principles remain largely unexplored in the context of electronic health records (EHRs) — a rich, sequential, and globally abundant data source that differs structurally from natural language. In this work, we present the first empirical investigation of scaling laws for EHR foundation models. By training transformer architectures on patient timeline data from the MIMIC-IV database across varying model sizes and compute budgets, we identify consistent scaling patterns, including parabolic IsoFLOPs curves and power-law relationships between compute, model parameters, data size, and clinical utility. These findings demonstrate that EHR models exhibit scaling behavior analogous to LLMs, offering predictive insights into resource-efficient training strategies. Our results lay the groundwork for developing powerful EHR foundation models capable of transforming clinical prediction tasks and advancing personalized healthcare.
\end{abstract}
\section{Introduction}

The discovery of scaling laws for large language models (LLMs) has significantly advanced the understanding of how model performance predictably improves as we scale up model size, dataset size, and computational resources for training~\citep{scalinglawsopenai,scalinglawschinchilla}. These scaling laws, initially characterized by training transformers on vast amounts of internet text data, have enabled the systematic development of increasingly powerful frontier models~\citep{gpt-3,gpt-4,gpt-4.5}.

Despite the explosion in data used to train LLMs, a substantial class of data remains largely untapped: electronic health records (EHRs). There are two primary barriers accounting for this underutilization. First, EHRs are proprietary and protected under strict privacy regulations, preventing open sharing and centralized access. Second, although EHRs contain text-based information, the longitudinal patient data comprising medical events and attributes are often standardized using specific coding systems, such as ICD codes~\citep{cms_icd10_2025} for diagnoses and ATC codes~\citep{who_atc_2025} for medications. This structured and codified information fundamentally differs from the natural language text typically processed by pretrained LLMs, presenting unique challenges for conventional LLMs.

Nevertheless, EHRs share critical similarities with natural language text that make them amenable to transformer-based modeling~\citep{transformer}. The events and attributes recorded in EHRs naturally form sequential timelines of patient encounters and treatments, analogous to word sequences in text data. Additionally, the volume at which EHRs are generated globally mirrors the scale of internet data, suggesting that foundation models trained on EHRs could benefit similarly from scaling laws.

Motivated by these similarities and opportunities, our work provides an initial exploration into scaling laws specifically for EHR foundation models. We train transformer models of varying scales on patient timeline data derived from the MIMIC-IV database~\citep{mimic-iv}, and systematically examine how model performance evolves with changes in model size, number of training tokens, and computational budgets. Our scaling-law experiments yield several key findings:

\begin{itemize}[leftmargin=1.5em]
    \item For a fixed computational budget (measured in floating-point operations, FLOPs), the IsoFLOPs curve — depicting the loss of EHR models relative to different model sizes — exhibits a clear parabolic relationship, consistent with scaling patterns observed in traditional LLMs. This curve enables estimation of the optimal model size that achieves minimal loss for a given computational budget.
    \item By tracing the loss minima of IsoFLOP curves across varying compute budgets, we derive empirical power-law relationships linking compute, optimal model size, and optimal number of training tokens. These power laws enable us to predict performance improvements as model and data scales increase. Moreover, they reveal predictable scaling behavior and underscore the importance of maintaining a careful balance between model size and data volume. 
    \item Zero-shot evaluation on downstream healthcare applications shows strong correlation between lower validation loss and better predictive performance, reinforcing the validation loss of EHR modeling as a guiding metric for EHR model development.
    \item Scaling model size lead to consistent improvements in clinical utility up to 28M parameters, beyond which performance begins to saturate due to insufficient training data. This plateau reflects the underfitting-overfitting tradeoff when models outscale the available token volume, and underscores the need for larger EHR datasets to sustain scaling gains.
\end{itemize}

This preliminary characterization of scaling laws for EHR foundation models provides predictive insights into the expected performance improvements achievable by scaling model size, expanding EHR training datasets, and increasing computational resources. Ultimately, these scaling laws lay critical groundwork for realizing transformative advancements — akin to the ``GPT moment'' — in predicting key clinical outcomes such as mortality risk, ICU admissions, and hospital stays, thus paving the way for significant improvements in healthcare delivery and patient outcomes.
\section{Methods}

\subsection{Problem Formulation}
Electronic health records (EHRs) consist of temporally ordered clinical events and attributes, forming a natural sequence akin to text in language modeling. Leveraging this sequential structure, we formulate the modeling of EHR data using an autoregressive framework similar to that employed in language models.

Let a patient timeline from EHRs be denoted as a variable-length sequence of medical tokens $\rvx = (x_1, x_2, \dots, x_N)$, where each token $x_i$ represents a distinct unit of clinical information, corresponding to diagnoses, medication administrations, hospital admissions, time intervals, or other meaningful elements from the patient's health trajectory (see \Cref{sec:patient_timeline_construct}).

Our goal is to model the joint probability distribution over the patient timeline $p(x_1,...,x_N)$, which we factorize as a product of conditional probabilities~\citep{bengio2003neural}:

$$
p(\rvx)=\prod_{i=1}^Np(x_i\mid x_1,...,x_{i-1})
$$

To parameterize these conditional probabilities, we employ a decoder-only transfomer neural network~\citep{transformer,gpt-1} with parameters $\theta$. The transformer model estimates each token’s likelihood given its preceding context as $f_\theta(x_1,...,x_i)=p_\theta(x_i\mid x_1,...,x_{i-1})$.  The model is trained via a standard next-token prediction objective, which maximizes the following log-likelihood:

$$
\mathcal{L}(\rvx)=\sum_{i=1}^N\log p_\theta(x_i\mid x_1,...,x_{i-1})
$$

During inference, given a partial patient timeline $(x_1,...,x_i)$, the model can autoregressively generate future tokens by sampling from the conditional distribution:

$$
(x_{i+1}, ..., x_N)\sim p_\theta(x_{i+1},...,x_{N}\mid x_1,...,x_i)=\prod_{j=i+1}^Np_\theta(x_j|x_1,...,x_{j-1})
$$

This formulation enables flexible modeling of patient trajectories, and supports sequence completion for large-scale training and inference of EHR foundation models.

\subsection{Database}

We choose MIMIC-IV~\citep{mimic-iv} as the source database for our experiments. MIMIC is a large, publicly available database of EHRs that provides comprehensive clinical data for research purposes. The dataset encompasses de-identified health records of over 200,000 adult patients who were admitted to the emergency department or an intensive care unit (ICU) at Beth Israel Deaconess Medical Center in Boston, Massachusetts, between 2008 and 2019. MIMIC-IV is designed to reflect real-world hospital operations, with data collected and timestamped continuously across a wide range of clinical encounters.

MIMIC-IV captures a rich array of data modalities, including:

\begin{itemize}[itemsep=1pt, topsep=1pt]
    \item \textbf{Diagnoses} recorded using ICD codes,
    \item \textbf{Medications} prescribed and administered during hospital stays,
    \item \textbf{Laboratory test results} and their associated values,
    \item \textbf{Procedures and clinical events}, such as surgeries or imaging,
    \item \textbf{Vital signs} and nursing assessments,
    \item \textbf{Demographics} and admission information,
    \item \textbf{Outcomes}, such as in-hospital and post-discharge mortality.
\end{itemize}

All data entries are localized in time, enabling the construction of detailed \emph{longitudinal patient timelines} — a critical requirement for sequential modeling using transformer-based architectures. The dataset's scope, scale, and structure make it particularly well-suited for investigating scaling laws in EHR foundation models. Its public availability also ensures that our findings are reproducible and extendable by the broader research community.

\subsection{Patient Timeline Construction}
\label{sec:patient_timeline_construct}

We employ the methodology and the codebase introduced by ETHOS~\citep{ethos} to construct patient timelines from the MIMIC-IV database, where chronologically ordered sequences of clinical events are transformed into discrete tokens, enabling autoregressive modeling over a patient’s longitudinal health trajectory.

\paragraph{ETHOS Approach}
To build these timelines, it extracts data from 12 core tables in MIMIC-IV, including \emph{diagnoses, laboratory test results, medications (converted to ATC codes~\citep{who_atc_2025}), procedures (ICD-10-PCS~\citep{cms_icd10_2025}), demographic attributes, ICU stays, DRG codes, SOFA scores}, and \emph{vital signs}. All events with timestamps are chronologically aligned based on the patient’s age at the time of the event, represented as floating-point values for temporal precision. Events lacking timestamps (e.g., demographic data) are encoded as static tokens and positioned at the beginning of each timeline to preserve causality.

Each clinical event is tokenized into 1 to 7 discrete tokens, designed to encode key semantic elements — such as event type, code hierarchy, and value category. For instance, hierarchical medical codes (e.g., ICD-10, ATC) are decomposed into multiple tokens that reflect their structural meaning. Numerical lab values and scores are tokenized using \textbf{quantile-based binning}, which transforms continuous measurements into discrete categorical representations that preserve clinical interpretability.

To represent temporal gaps between events, \textbf{time-interval tokens} are inserted throughout the sequence. Thirteen interval classes (e.g., 5 minutes, 6 months) are used to discretize inter-event times, ensuring that the model captures temporal dynamics. No interval token is added if the elapsed time is less than 5 minutes; long gaps are approximated using repeated interval tokens (e.g., three 6-month tokens to represent 1.4 years).

Additionally, \textbf{age and timeline start year} are encoded as coarse 5-year interval tokens to provide temporal context while respecting MIMIC’s date obfuscation. This approach reflects ETHOS’s assumption that both patient health trajectories and medical practice patterns evolve gradually over time, obviating the need for fine-grained temporal resolution.

We also adopt ETHOS’s constraints for sequential modeling, ensuring that only information available at or before a given time point is used during training and inference. For example, DRG codes (assigned at discharge) and SOFA scores (initially assessed in the ICU) are placed in the timeline only after related admission or discharge events, and are excluded from input during prediction tasks to prevent information leakage.

\paragraph{Our Modification}
One key distinction between our approach and that of ETHOS lies in how training batches of patient timelines are constructed. In ETHOS, all patient timelines are first concatenated into a single, continuous sequence, which is then divided into fixed-length chunks. Training batches are randomly sampled from these chunks. However, this strategy introduces a major limitation: individual chunks may span across multiple patient timelines, leading to situations where tokens from one patient's history are used to predict the next token of another patient — an outcome that violates the temporal and clinical independence of patient records.

To address this issue, we adopt a boundary-preserving batching strategy. Specifically, we segment each patient timeline independently, ensuring that all training sequences remain fully contained within a single patient's trajectory. This design respects the natural boundaries of patient histories and prevents information leakage across timelines.

In total, we construct patient timelines for 267,773 patients from MIMIC-IV. We follow ETHOS using data from 90\% patients for training and validation and the remaining 10\% for test. The statistics of training and test splits are reported in \Cref{tab:data_stats}.

\begin{table}[!ht]
\centering
\small
\begin{tabular}{@{}lrrr@{}}
\toprule
 & \textbf{Train} & \textbf{Validation} & \textbf{Test} \\ \midrule
patients & 217,466 & 23,549 & 26,758 \\
\multicolumn{4}{@{}l}{patient timeline length (measured by tokens)} \\
\quad mean/std & 1,227/3,363 & 1259/3,546 & 1217/3,372 \\
\quad min/max & 2/217,971 & 2/190,131 & 2/87,633 \\
\quad Q1/Q2/Q3 quartiles & 96/308/1,028 & 95/313/1,062 & 96/313/1,009 \\
total timeline tokens & 266,766,449 & 29,651,324 & 32,565,814 \\
\multicolumn{4}{@{}l}{training example length (measured by tokens)} \\
\quad mean/std & 962/810 & 979/814 & 958/807 \\
\quad min/max & 37/2,048 & 37/2,048 & 37/2,048 \\
\quad Q1/Q2/Q3 quartiles & 184/660/2,048 & 189/686/2,048 & 186/654/2,048 \\
total training examples & 278,483 & 30,419 & 34,142 \\ 
total trainable tokens & 267,942,937 & 29,779,264 & 32,709,535 \\
\bottomrule
\end{tabular}
\caption{Statistics of constructed patient timelines for training, validation, and test.}
\label{tab:data_stats}
\end{table}

\subsection{Model Implementation}

We adopt the Llama architecture~\citep{llama1,llama3} as implemented in the HuggingFace Transformers library~\citep{huggingface}. Llama represents a modern, optimized variant of the standard decoder-only transformer, incorporating several enhancements that improve training stability and efficiency. These include pre-layer normalization~\citep{gpt-3}, SwiGLU activations~\citep{shazeer2020gluvariantsimprovetransformer}, rotary positional embeddings~\citep{su2023roformerenhancedtransformerrotary}, and grouped query attention~\citep{ainslie2023gqatraininggeneralizedmultiquery} — all of which have been shown to contribute to strong performance in large-scale language modeling.

It is important to note that we \emph{do not use any pretrained weights} from the original Llama models. Instead, we use the \emph{Llama architecture purely as a backbone}, initializing all models from scratch with random weights. This design choice ensures a controlled environment for exploring scaling laws, where performance improvements are driven solely by model scale and training dynamics, rather than transfer learning from general-domain textual corpora.
Exploring initialization with pretrained Llama weights is left as an avenue for future work.
\section{Experiment Setup}
\label{sec:setup}

To investigate scaling laws for EHR foundation models, we design a series of controlled experiments that systematically vary model size and training compute. In this section, we detail the configurations used in our study, including model sizes, compute budgets, and training hyperparameters. These design choices enable a consistent and reproducible framework for analyzing how performance scales with model capacity and data.

\noindent\textbf{Model Sizes}
To systematically examine scale effects, we construct a family of transformer-based models spanning a wide range of parameter counts -- from sub-10M to nearly 1B. 
Our central model contains roughly 50–60 million parameters, chosen based on its strong trade-off between compute efficiency and downstream performance.
Model variants are generated through controlled adjustments to core architectural parameters, including the depth, width, and attention configuration. These variants maintain adherence to transformer design principles and follow heuristics inspired by recent open-source models.

\noindent\textbf{Hyperparameters}
Our training setup reflects a convergence of best practices reported across recent foundational studies in scaling laws. While an exhaustive sweep across all training hyperparameters would offer fine-grained optimization, we adopt a pragmatic strategy: selecting widely-used defaults shown to work robustly across model scales.

We employ a standard optimizer and schedule configuration, with a cosine learning rate decay and warmup, modest weight decay, and stable training enabled by modern precision formats. We maintain a fixed context length and token budget per batch across models, ensuring comparability in training dynamics. 
The selected configuration prioritizes reproducibility and computational tractability, offering a stable foundation for assessing scaling behavior without the confounding influence of aggressive per-model tuning.

\noindent\textbf{Compute Budget}
To quantify the compute budget in our experiments, we adopt floating-point operations (FLOPs) as the primary metric. Following the calculation outlined in the Chinchilla paper, we estimate the total training FLOPs by first computing the cost of a forward pass as:
$$
\text{FLOPs}_{\text{forward}} \approx \text{num\_layers} \times (\text{total\_attention} + \text{dense\_block})
$$
The backward pass is assumed to be approximately twice as expensive as the forward pass, leading to a total training cost of roughly $3\times$ forward FLOPs per training example.

To ensure consistency, we also compare this FLOPs estimation method with that used in the PaLM paper. Our analysis shows that the difference between the two approaches is small and does not significantly impact the resulting compute estimates.

In our experiments, we apply the Chinchilla-based calculation to estimate the total FLOPs required to train a model of a given size (as specified in Table 1). Based on this, we determine the number of training tokens that can be processed under a fixed compute budget.

\section{Results}
In this section, we present empirical findings from our scaling experiments on EHR foundation models. 
We begin by analyzing IsoFLOP profiles to quantify the relationship between compute, model size, and validation loss.
We then evaluate how these trends translate to downstream clinical utility through zero-shot performance on practical healthcare prediction tasks.
Together, these results establish foundational insights into how EHR models benefit from scaling.

\subsection{IsoFLOP Profiles}
\label{sec:isoflop}

To analyze the relationship between compute, model size, and performance, we follow the ``Approach 2'' from \cite{scalinglawschinchilla} to construct IsoFLOP profiles.
In this setup, we fix the total training FLOPs across a range of budgets (from $5\times10^4$ to $8\times10^5$ TFLOPs),
and systematically vary the model size.
For each configuration, the number of training tokens is determined the model's FLOPs consumption, following the methodology described in \Cref{sec:setup}.
The core question we aim to answer is: \emph{Given a fixed compute budget, what is the model size that minimizes validation loss?}

\begin{figure}[!ht]
    \centering
    \includegraphics[width=1.0\textwidth]
    {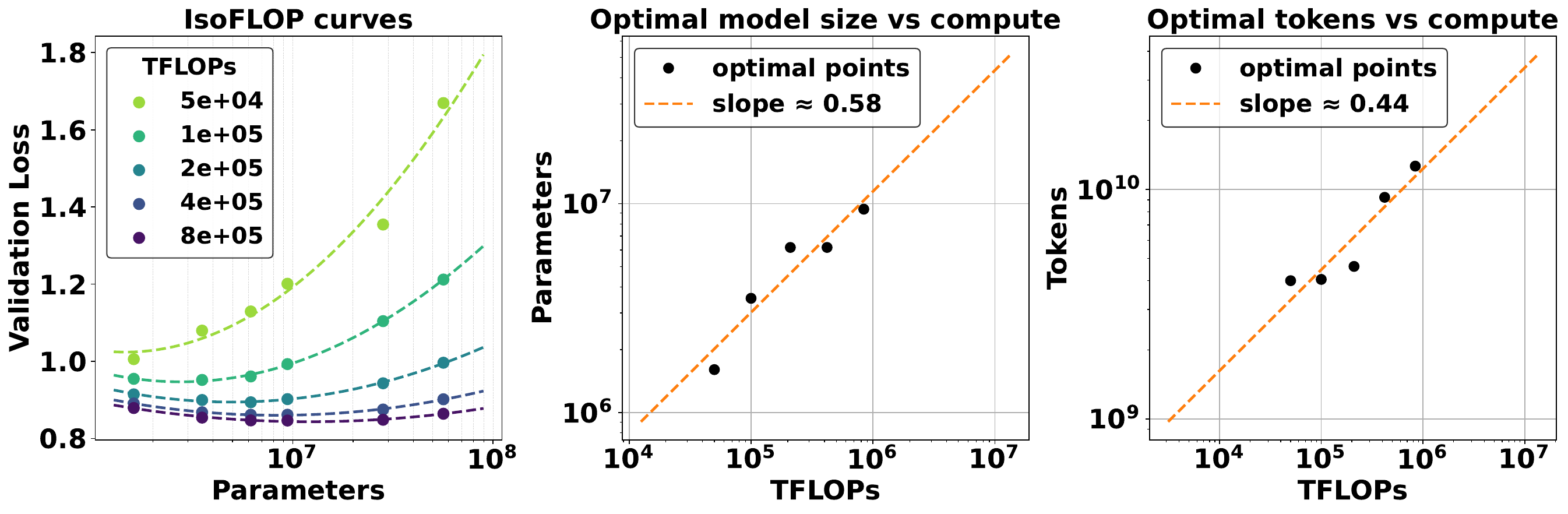}
    \caption{\textbf{IsoFLOP curves.} \emph{Left}: For various model sizes, we select the number of training tokens such that the total training FLOPs remain constant. A distinct loss valley emerges, indicating an optimal model size under each compute budget.
    \emph{Center and Right:} Using the positions of these minima, we extrapolate optimal model size and number of training tokens for larger FLOP budgets.}
    \label{fig:iso_flops}
\end{figure}

\Cref{fig:iso_flops} (left) illustrates the IsoFLOP profiles.
For each FLOP budget, we train all model configurations described in \Cref{sec:setup},
then retain and plot the six models with the lowest validation loss against their corresponding parameter counts.
This selection ensures a diverse yet representative sampling of model sizes while filtering out clear outliers with substantially higher losses.
We observe that for each compute budget, the loss exhibits a well-defined \textbf{U-shaped profile} when plotted against model size.
We fit a parabolic curve to each IsoFLOP profile, which allows us to precisely estimate the loss-optimal model size, i.e., the point at which training efficiency is maximized for the given compute.

Building on these optimal points, we then analyze how the loss-optimal model size and number of training tokens scale as a function of the compute budget. The center and right panels of \Cref{fig:iso_flops} show that both quantities follow a clear power-law relationship, indicating that scaling laws also hold for EHR foundation models. Specifically, we fit the following relationships:
$N_\textrm{opt}\propto C^a$ and $D_\textrm{opt}\propto C^b$, where $N$ is the number of model parameters, $D$ is the number of training tokens, and $C$ is the compute budget in FLOPs.
From our fitted curves, we find that $a\approx0.58$ and $b\approx0.44$.
These exponents differ from those reported in \cite{scalinglawschinchilla} for LLMs, suggesting fundamental differences in scaling dynamics between EHR data and natural language. We hypothesize two potential reasons:

\begin{itemize}[itemsep=1pt, topsep=1pt,leftmargin=1.5em]
    \item \textbf{Modality mismatch:} Medical events and attributes in EHRs, though sequential, are semantically and structurally different from natural language.
    \item \textbf{Data scale:} The total number of training tokens available from EHRs for our experiments is considerably smaller than the scale typically used in LLM studies, which may reduce the precision and stability of exponent estimation from the IsoFLOP profiles.
\end{itemize}

These findings represent the first empirical scaling profiles for EHR foundation models and offer guidance for optimal resource allocation in future model development.

\subsection{Scaling Laws on Healthcare Applications}

Our formulation -- modeling patient timelines autoregressively -- enables flexible, zero-shot inference for a range of clinical prediction tasks.
Once trained, the models can simulate future patient trajectories conditioned on partial history, without requiring additional task-specific fine-tuning. 
This capability allows us to study the clinical utility of scaling EHR foundation models.
Specifically, we systematically evaluate the relationship between downstream healthcare task performance, validation loss, and model size.

Following the ETHOS evaluation protocol \citep{ethos}, we assess zero-shot performance on two representative prediction tasks using 10\% of the patients in MIMIC-IV, held out as a test set:

\begin{itemize}[itemsep=1pt, topsep=1pt,leftmargin=1.5em]
\item\textbf{ICU Mortality}
We estimate the probability of in-ICU mortality at the time of ICU admission.
For each patient, we begin generation from the ICU admission token and let the model to generates one token at a time until it emits either a discharge or death token. This process is repeated 20 times, and the probability of ICU mortality is estimated as $N/20$, where $N$ is the number of simulations that terminate with a death token.
\item\noindent\textbf{30-Day Inpatient Readmission} 
To estimate the probability of inpatient readmission within 30 days of discharge, generation begins at the discharge token from the last inpatient stay. The model continues generating tokens until it produces either a new admission, death, or until the cumulative time-interval tokens exceed 30 days. This simulation is repeated 20 times, and the readmission probability is computed as $M/20$, where $M$ is the number of sequences that terminate with a new admission token.
\end{itemize}

\noindent\textbf{Inference Procedure}
In both tasks, the model conditions on each patient’s historical timeline up to the relevant decision point: either the ICU admission or discharge. 
Up to 2048 tokens from the patient's history are included (or the full timeline if shorter) to initiate the inference. 
The model then generates one token at a time through the following steps: (1) The model computes probabilities for all potential tokens; (2) A new token is sampled stochastically based on these probabilities; (3) The token is appended to the sequence, and the earliest token is removed to maintain a 2048-token context window; (4) Generation continues until a predefined stopping condition is met (e.g., death token, discharge, or time limit exceeded).
The stochastic sampling allows us to simulate multiple plausible future trajectories, capturing predictive uncertainty inherent in clinical forecasting.

\noindent\textbf{Evaluation Metrics}
We evaluate model performance using Receiver Operating Characteristic (ROC) curves. ROC curves are fitted using Gaussian models with unequal variances for binary classification, and Area Under the Curve (AUC) values are computed along with 95\% confidence intervals via bootstrapping. We adopt the open-source evaluation code provided by ETHOS at \href{http://github.com/ipolharvard/ethos-paper}{github.com/ipolharvard/ethos-paper}.

\noindent\textbf{Evaluated Models}
We evaluate two groups of models:
(1) \textbf{IsoFLOP-optimal models} with parameter counts ranging from 1M to 9M, identified based on the lowest validation loss under fixed compute budgets (see \Cref{sec:isoflop}).
(2) \textbf{Larger-scale models}, ranging from 28M to 982M parameters, which are not optimal under the fixed budgets but are trained until early stopping (i.e., when validation loss begins to increase, indicating overfitting). For these, we report results from the checkpoints achieving the lowest validation loss.

\begin{figure}[!ht]
    \centering
    \includegraphics[width=1.0\textwidth]
    {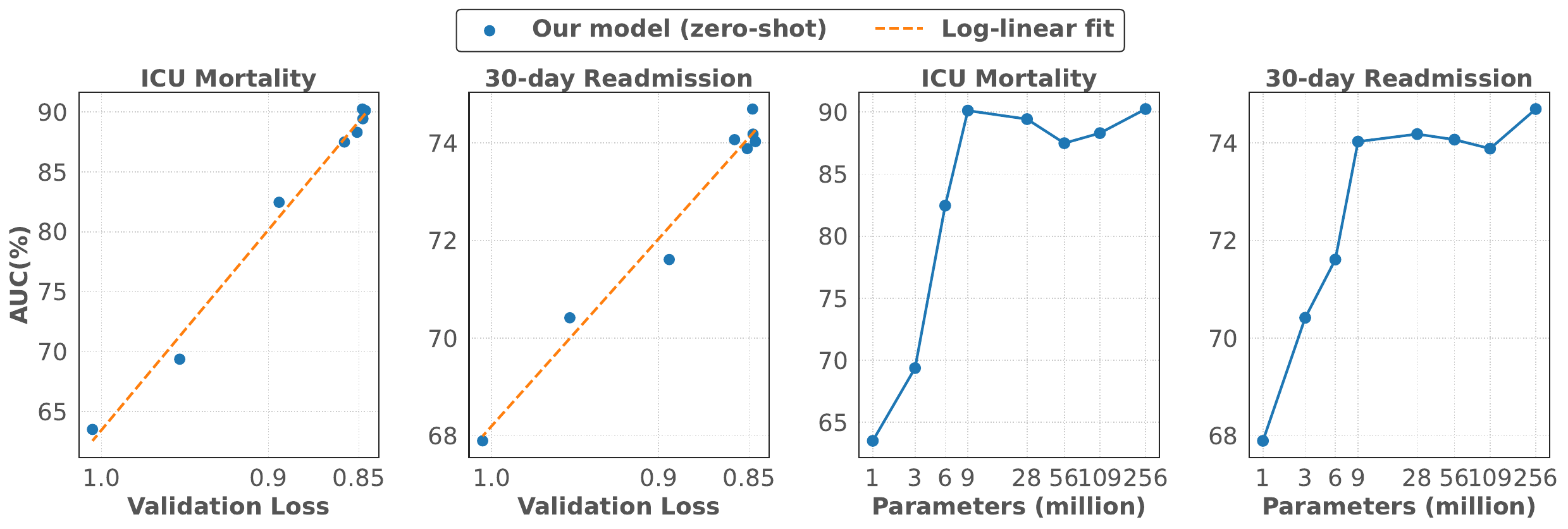}
    \caption{Zero-shot performance on downstream healthcare tasks. \emph{Left:} Task performance correlate strongly with the model's validation log-loss. \emph{Right:} Task performance improves with increasing model size (measured by parameters) but saturates beyond 28M parameters, due to the limited number of training tokens.}
    \label{fig:downstream_eval}
\end{figure}

\noindent\textbf{Results and Analysis}
\Cref{fig:downstream_eval} illustrates the relationship between validation loss, model size, and zero-shot performance on ICU mortality and 30 day inpatient readmission. 
In the left two panels, we observe a strong linear correlation between the log of validation loss and task performance.
For instance, as the validation loss drops from 1.0 to 0.85, ROC AUC for ICU mortality improves from 0.68 to 0.75, while PR AUC increases from 0.39 to 0.50. A similar trend is observed for 30-day inpatient readmission.
This demonstrates that \emph{validation loss serves as a reliable proxy} for downstream task performance, reinforcing its value as a key metric for model selection and scaling-law analysis.

The right two panels of \Cref{fig:downstream_eval} show how performance scales with model size. From 1M to 9M parameters, both tasks exhibit consistent, log-linear improvements. However, beyond 28M parameters, performance gains begin to plateau, with diminishing returns observed across all metrics.
We hypothesize that this saturation arises from data scarcity. MIMIC-IV contains approximately $2.67\times10^8$ training tokens (\Cref{tab:data_stats}).
According to the power-law scaling trends identified in \Cref{fig:iso_flops}, the optimal number of training tokens for models larger than 28M parameters exceeds $10^{10}$ — over two orders of magnitude greater than what the dataset provides.
Attempting to scale model size without proportionally increasing the dataset leads to overfitting, as larger models memorize the training data rather than generalizing. This is reflected in the cluster of large models in the left panels of \Cref{fig:downstream_eval} that converge to similar validation losses ($\sim0.85$), mirroring the performance of the smaller, well-calibrated 9M models.
The ROC curves of these models are provided in \Cref{sec:appendix}.

These findings suggest that continued log-linear scaling in downstream task performance is possible, but only when accompanied by proportional increases in training data volume. Extending this trend with larger EHR datasets remains an important future direction.

\section{Related Work}
\paragraph{Scaling Laws}
The discovery and systematic study of scaling laws have profoundly shaped the development of large language models (LLMs).
\cite{scalinglawsopenai} first demonstrated that validation loss follows a power-law decay with increasing model size, dataset size, and compute.
\cite{henighan2020scaling} extended this to multimodal settings, confirming the generality of these trends. 
In parallel, \cite{gpt-3} showed that extreme scale alone enables emergent few-shot capabilities in GPT-3.
Further refinements to scaling strategies were proposed by \cite{scalinglawschinchilla}, who observed that smaller models trained on more data (e.g., Chinchilla) can outperform much larger under-trained models.
\cite{chowdhery2022palmscalinglanguagemodeling} scaled further with PaLM (540B), which displayed state-of-the-art performance and emergent reasoning skills.
Collectively, these works establish predictable scaling behaviors and highlight how careful balance between model size and data governs efficiency and generalization.
This body of work directly motivates our exploration of scaling laws for EHR foundation models.

\paragraph{EHR Foundation Models}
Inspired by the success of LLMs, foundation models for  electronic health records (EHRs) aim to learn general-purpose representations from patient timelines.
Early efforts like BEHRT~\citep{li2020behrt} and Med-BERT~\citep{rasmy2021med} applied BERT modeling~\citep{bert} to structured records from a large number of patients, improving disease prediction.
ETHOS~\citep{ethos} introduced generative transformer modeling over tokenized patient events, while EventStreamGPT~\citep{mcdermott2023event} and TransformEHR~\citep{yang2023transformehr} refined autoregressive and seq-to-seq approaches for dynamic patient timelines.
\cite{kraljevic2024foresight} integrated unstructured clinical text with structured data, using NLP tools to convert free-text notes into coded concepts.
\cite{wornow2023shaky} conducted a comprehensive review of 84 EHR-based foundation models and observed that many have been trained on relatively narrow datasets and evaluated on proxy tasks that may not reflect real clinical utility.
Building upon these prior efforts, our work systematically examines the scaling behavior of EHR foundation models, quantifying how model performance improves with scale under fixed compute budgets.
\section{Conclusion}

We present an initial systematic exploration of scaling laws for EHR foundation models, leveraging transformer architectures trained on patient timelines constructed from the MIMIC-IV database. By tracing IsoFLOP profiles across varying compute budgets, we uncover empirical power-law relationships between model size, training tokens, and compute—highlighting the existence of predictable scaling behavior in the healthcare domain. Our findings demonstrate that validation loss serves as a strong proxy for downstream clinical utility, with performance on critical tasks like ICU mortality and 30-day hospital readmission improving steadily with scale up to 28 million parameters. Beyond this point, performance gains saturate due to data limitations, underscoring the importance of scaling both model capacity and EHR datasets in tandem. This study provides practical guidance for training compute-optimal models on structured medical data and paves the way for future work on scaling EHR foundation models to unlock their full potential in real-world clinical applications.

\bibliography{references}
\bibliographystyle{colm2025_conference}

\newpage
\appendix
\section{Appendix}
\label{sec:appendix}

\begin{figure}[!ht]
    \centering
    \includegraphics[width=1.0\textwidth]{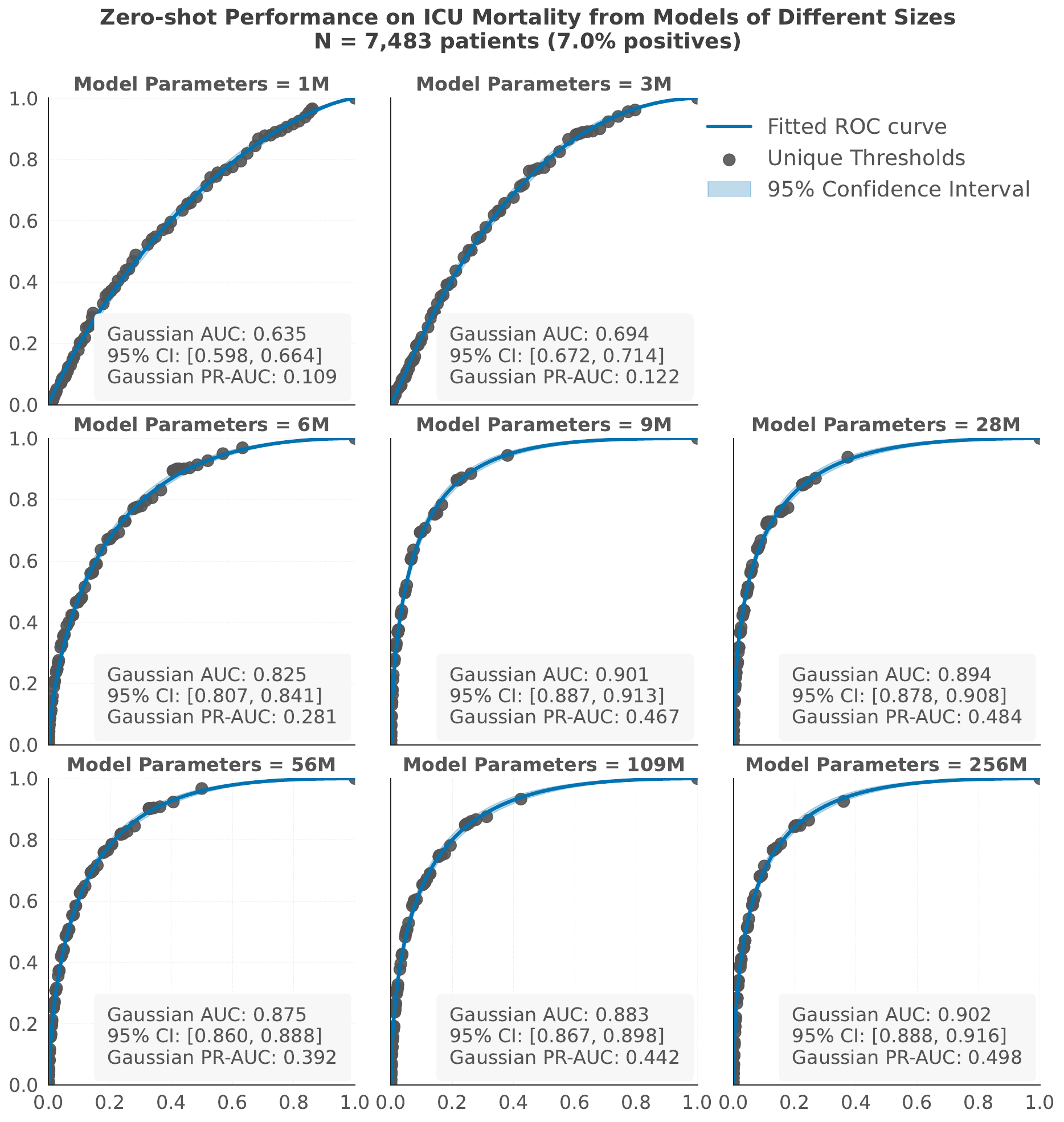}
    \caption{Zero-shot performance of models of different sizes on ICU mortality.}
    \label{fig:icu_mortality}
\end{figure}

\begin{figure}[!ht]
    \centering
    \includegraphics[width=1.0\textwidth]{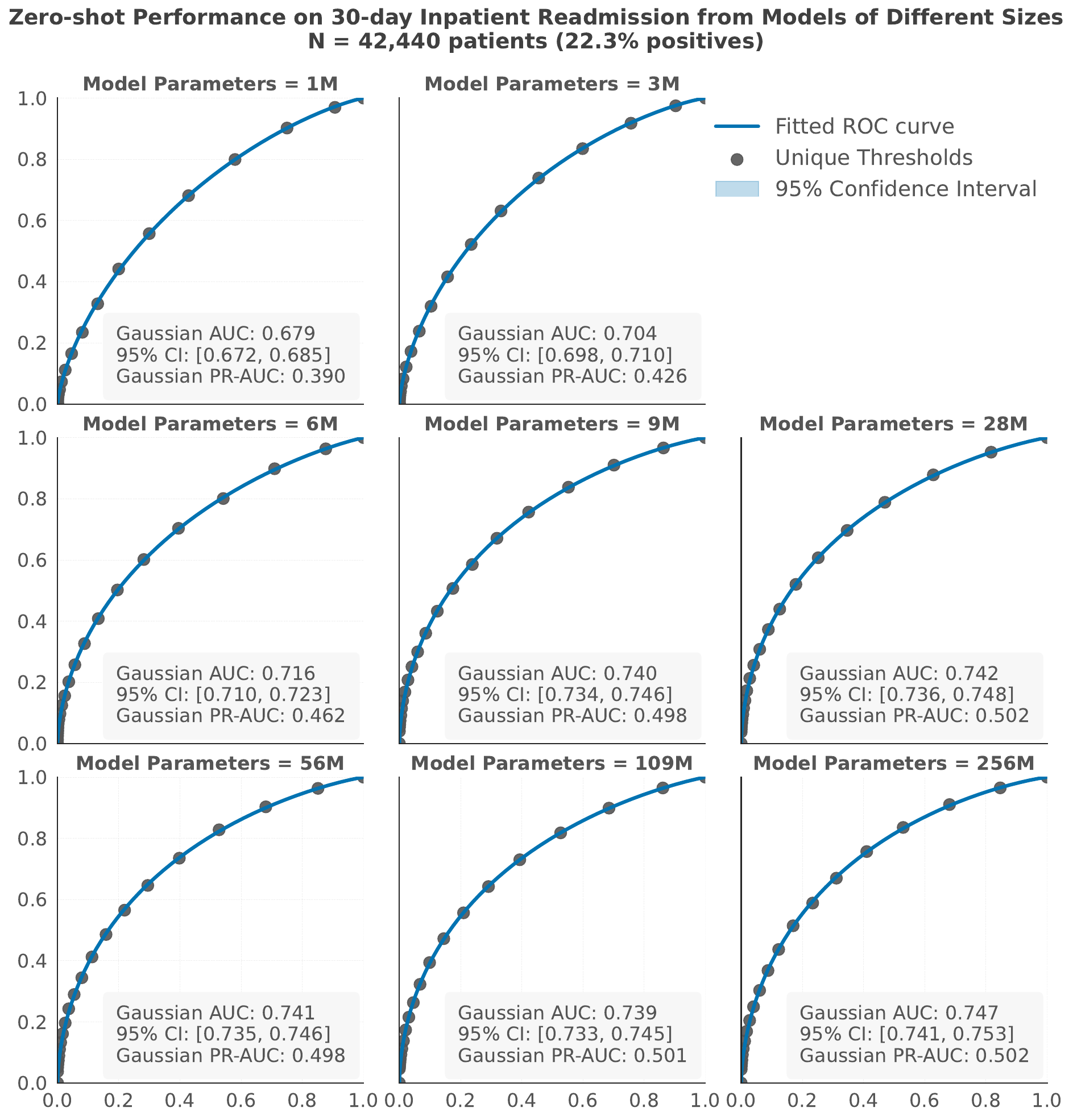}
    \caption{Zero-shot performance of models of different sizes on 30-day hospital readmission.}
    \label{fig:hospital_readmission}
\end{figure}

\end{document}